\documentclass[11pt,a4paper]{article}
\usepackage[utf8]{inputenc}
\usepackage[hyperref]{emnlp-ijcnlp-2019}
\usepackage{times}
\usepackage{latexsym}
\usepackage{verbatim}
\usepackage{amsmath}
\usepackage{amssymb}
\usepackage{float}
\usepackage{placeins}
\usepackage[OT1]{fontenc}
\usepackage{graphicx}
\usepackage{algorithm}
\usepackage{hyperref}
\usepackage[noend]{algpseudocode}
\usepackage[small,compact]{titlesec}
\usepackage[font=small,aboveskip=4pt,belowskip=0pt]{caption}
\usepackage{enumitem}
\usepackage[nopar]{lipsum}
\setdescription{noitemsep,topsep=0pt}
\graphicspath{ {figures/} }

\DeclareMathOperator{\softmax}{softmax}
\DeclareMathOperator{\argmax}{argmax}
\DeclareMathOperator{\lstm}{LSTM}
\DeclareMathOperator{\affine}{affine}
\DeclareMathOperator{\enc}{enc}
\DeclareMathOperator{\dec}{dec}

\DeclareMathOperator{\pr}{P}
\DeclareMathOperator{\lex}{Lex}
\makeatletter
\def\BState{\State\hskip-\ALG@thistlm}
\makeatother
\algnewcommand{\LeftComment}[1]{\Statex \(\triangleright\) #1}

\usepackage{url}

\aclfinalcopy 

\title{Pointer-based Fusion of Bilingual Lexicons into Neural Machine
    Translation}

\author{Jetic G\=u, Hassan S. Shavarani, and Anoop Sarkar \\
  Simon Fraser University \\
  8888 University Drive \\
  Burnaby, BC, Canada \\
  {\tt \{jeticg, sshavarani, anoop\}@sfu.ca} \\}

\date{}

\begin{document}
\maketitle
\begin{abstract}
  Neural machine translation (NMT) systems require large amounts of high
  quality in-domain parallel corpora for training.
  State-of-the-art NMT systems still face challenges related to out of
  vocabulary words and dealing with low-resource language pairs.
  In this paper, we propose and compare several models for fusion of bilingual
  lexicons with an end-to-end trained sequence-to-sequence model for machine
  translation.
  The result is a fusion model with two information sources for the
  decoder: a neural conditional language model and a bilingual lexicon.
  This fusion model learns how to combine both sources of information in order
  to produce higher quality translation output.
  Our experiments show that our proposed models work well in relatively
  low-resource scenarios, and also effectively reduce the parameter size and
  training cost for NMT without sacrificing performance.
\end{abstract}

\section{Introduction}
Recent years have seen rapid progress in neural machine translation (NMT)
research.
While fully neural systems with attention mechanism \citep{CL2015008,
CL2015007, CL2017244} achieve the state-of-the-art performance on several
benchmarks, digesting large source and target side vocabulary sets remains a
major challenge.
Neural models rely on either word embedding matrices or character-level
encoders, which constrains the ability of the models of handling unseen or rare
words.
For word-embedding-based NMT, models commonly treat these words (including
named entities) as Out-of-Vocabulary (OOV) and replace with a unified
\texttt{UNK} token.
Meanwhile, language naturally evolves through time.
Adaptation to such evolution for a neural model can mean laborious gathering of
new data and fine-tuning millions of parameters.
Sensitive domains such as Biomedical also require high precision terminology
translation that is hard to guarantee in neural space.

An intuitive way to tackle these issue is to utilise a copy mechanism or
separate transliteration mechanism.
\citet{CL2015009} propose \texttt{UNK} replacement based on Pointer Network
\citep{ML2015001}.
After neural decoding, the model replaces \texttt{UNK} tokens by looking up the
source word with highest attention score and copying it over.
\citet{CL2016032} propose to use pointer generator to determine
when to copy, and a separate pointer scoring layer instead of attention scores.
\citet{CL2016359, CL2017343} focus on named entities by adding a separate
char-level transliteration model.
This line of work however, does not cover more general lexical categories.

In this paper, we propose a novel way to combine prior knowledge from a
dictionary-like bilingual lexicon with a fully neural translation model.
We extend pointer-based copy with a translation table (dictionary) and its
entry-level features.
Such knowledge are all automatically generated from the training data, although
we also demonstrate that the inclusion of external knowledge is also possible
and beneficial.
The result is a fusion model with a contextual neural language model
combined with a bilingual lexicon\footnote{Code:
\url{https://github.com/jeticg/Code013-CL-LexPG}}.
Our experimental results show that our fusion model retains translation
quality when the number of its neural parameters is dramatically reduced to
even a fourth of the original size.

\section{Related Work}
\label{sec:Related}
The subject of utilising external lexical knowledge has been an interesting
topic for NMT.

\citet{CL2015014, CL2016242} use a dictionary to limit the target search space,
improving inference efficiency.
\citet{CL2017160, CL2018455} enforce lexical constraints in beam search.
\citet{CL2018456} further uses finite-state acceptors in beam search to ensure
decoding conforms to multiple lexical constraints.
The decoder here is influenced by an external scoring function that is
knowledge-aware, improving interpretability and flexibility.
Although as \citet{CL2017336} claim, the fact that the neural model is unaware
of the lexical constraints may lead to sub-optimal translation.

Alternatively, \citet{CL2016356} investigate using the alignment scores as
regularisers.
The distilled lexical knowledge is incorporated into the probability
distribution calculation as a bias term.

A separate line of work focuses on data manipulation.
\citet{CL2015081} use the bilingual anthology to filter out domain-irrelevant
samples for training.
\citet{CL2019033} also perform data synthesis by replacing the source rare
words with their translations from a bilingual lexicon.

\section{Neural Machine Translation and Copy Mechanism}
\label{sec:NMT}
Machine translation are models for which the input token sequence
$X = (x_0, x_1, ..., x_{n-1})$ in language 1 (source) is transformed into its
semantically equivalent output sequence $Y = (y_0, y_1, ..., y_{m-1})$ in a
second language (target).
Neural machine translation models are ones in which the conditional language
model $\pr(Y|X)$ is computed using neural networks.

Common NMT models use encoder-decoder architecture \citep{CL2014005}.
The encoder transforms the input into hidden representation space, to be taken
by the decoder as input to generate the final output (distribution).
The baseline model we use in this paper is a sequence-to-sequence model under
such architecture with attention \citep{CL2015008, CL2015007}.
The encoder and decoder are both LSTM units \citep{ML1997001}, more
specifically the encoder is a bidirectional LSTM that processes the input
sequence from both directions (Equation \ref{equ:NMT:enc}).

\begin{equation}
    \begin{aligned}
    \label{equ:NMT:enc}
    \overrightarrow{h}^{\enc}_i &= \overrightarrow{\lstm}
        (\overrightarrow{h}^{\enc}_{i-1}, x_i) \\
    \overleftarrow{h}^{\enc}_i &= \overleftarrow{\lstm}
        (\overleftarrow{h}^{\enc}_{i+1}, x_i) \\
    h^{\enc}_i &= [\overrightarrow{h}^{\enc}_i; \overleftarrow{h}^{\enc}_i]
    \end{aligned}
\end{equation}

Then, at each step $t$, attention mechanism takes as input the decoder state
$h^{\dec}_t$ and all encoder states to generate context vector $c_t$
(Equation~\ref{equ:NMT:att}).

\begin{equation}
    \begin{aligned}
    \label{equ:NMT:att}
    \mathrm{score}_{ti} &=
        V \times \tanh(W \times h^{\dec}_t + U \times h^{\enc}_i) \\
    \alpha_t &= \softmax(\mathrm{score}_t) \\
    c_t &= \sum_i\alpha_{ti}h^{\enc}_i\\
    \end{aligned}
\end{equation}

Finally, $c_t$ and $h^{\dec}_t$ are combined to calculate the probability
distribution over the neural lexicon (Equation~\ref{equ:NMT:base_dec}).
After that, decoding continues by feeding both representations and the
generated word's embedding back into the decoder, until a end-of-sequence
token is omitted.

\begin{equation}
    \begin{aligned}
    \label{equ:NMT:base_dec}
    h^{\dec}_t &= \lstm^{\dec}(\hat{h}^{\dec}_{t-1}, y_{t-1}) \\
    \hat{h}^{\dec}_{t} &= \tanh(C \times [c_t; h^{\dec}_t]) \\
    \pr_{\dec}(y_t = w|y_{<t}, X) &= \softmax(\affine(\hat{h}^{\dec}_t))(w)
    \end{aligned}
\end{equation}

\citet{CL2016032, CL2016171, CL2017118} proposes pointer generator (PG), which
is essentially a switch that determines at each step $t$, whether to use the
neural decoder's output or to copy.
We formulate PG using an activation unit $p_{gen}$
(Equation~\ref{equ:NMT:copyPG}, the probability of using neural generation).

\begin{equation}
    \begin{aligned}
    \label{equ:NMT:copyPG}
    p_{gen} &= \sigma (w_{gen}^Tc_t + b_{gen}) \\
    \pr_{\mathrm{CopyPG}}(y_t = w &| y_{<t}, X) = \\
        p_{gen} &\times \pr_{\dec}(y_t = w | y_{<t}, X) + \\
        (1 - p_{gen}) &\times \sum_{x_i = w} \alpha_i  \\
    \end{aligned}
\end{equation}

During inference, when $p_{gen}$ is bigger than a certain threshold (we find
using extensive experimentation that 0.5 works the best) the model uses
the neural decoder's output, otherwise the copy mechanism is triggered.
The word copied from source is $\argmax_i \alpha_i$ entry $x_i$.

\section{Dictionary Fusion}
\label{sec:Lex}
\citep{CL2015009, CL2016032} shows copy mechanism can improve translation
quality, without a dictionary it only covers languages with significant
lexical overlap.
To put this into perspective, IWSLT2017 German-English train-set has 17006
words that occur on both sides, that is 13\% of the German lexicon and
33\% of the English lexicon.
This convenient use of the copy mechanism cannot be extended to languages with
different alphabet or writing systems.
Composition of bilingual word-level lexicons with the decoder is a non-trivial
task, leading to the lexical information being largely ignored in fully neural
MT systems: once the words are mapped or encoded into vectors, lexical level
similarity and relation across sides are no longer of any concern.

Another problem is one of rare words.
Synthesising datasets with desirable rare-word distribution is also
non-trivial and much harder than identifying and translating rare words in
isolation.
Storing rare word translations in bilingual dictionary is a convenient
way to deal with this issue but is largely ignored in contemporary NMT systems.

We propose to utilise such information by implementing a fusion model that
combines an encoder-decoder model with automatically extracted
dictionary-like bilingual lexicon.
Such lexicon is obtained using an off-the-shelf HMM-based Word Aligner.
Word alignment is commonly used in statistical MT pipeline \citep{CL2003003}
prior to the construction of phrase tables.
It learns word-level mapping using unsupervised learning on parallel corpora,
which we take as a dictionary alternative, though the utilisation of a real
dictionary is also possible.

We propose several ways of using such information and compare their
differences:
\begin{itemize}
  \item \textbf{LexPN}: utilise attention scores to replace \texttt{UNK}
        tokens \citep{CL2015009};
  \item \textbf{LexPG}: Pointer Generator \citep{CL2016032, CL2017118} with
        attention scores and language model scoring;
  \item \textbf{LexPG+S}: LexPG with separate pointer scoring layer;
  \item \textbf{LexPG+F}: LexPG with dictionary entry features;
  \item \textbf{LexPG+S+F}: Pointer-based fusion with separate pointer scoring
        layer and dictionary entry features.
\end{itemize}

\subsection{LexPN: Dictionary Fusion with Pointer Network}
\label{ssec:Lex:LexPN}
Pointer network uses attention scores as positional pointer to select tokens
from the input sequence.
\citet{CL2015008} show that attention mechanism can be used to indicate which
part of the input sequence was paid prominent attention to, and argues that
such information resembles word alignment.
\citet{CL2015009} show that translation quality can be improved by using
attention score ranking as copy index ($\alpha$ in Equation~\ref{equ:NMT:att}).
We extend their approach by not only copying from the source side, but when the
pointed source entry exists in the dictionary, take its translation as output.

LexPN model extends a Seq2Seq baseline by modifying the decoding process.
The end result is a translation model with 2 separate components: a neural
conditional generative language model (Equation~\ref{equ:NMT:base_dec}, a
standard Seq2Seq decoder) in which words are mapped to vectors using the
embedding matrix (the neural lexicon $\mathrm{nLex}$); and a symbolic
bilingual lexicon (dictionary) in which humanly interpretable word-level
translation pairs are stored with occurrence frequency obtained with the
alignment model.
During inference time, when the neural conditional LM generates an \texttt{UNK}
token, dictionary lookup is triggered by looking at attention scores to
determine from which source word to perform translation.
The chosen translation is then fed back into the neural decoder if it exists in
the neural lexicon, and if not the embedding of \texttt{UNK}.

It is important to note that the neural and the symbolic lexicon may cover
different words.
Ideally, it should be a ``contained'' relation, since the symbolic lexicon
(dictionary) is extensible at run-time, as opposed to the neural embedding
lookup table which is fixed to a set vocabulary size and in practice the size
is capped by word frequencies or using byte-pair encoding or similar criteria.
This problem may be avoidable by using character-level or sub-word level tokens
for some language pairs \citep{CL2016179}, but for newly invented words and
proper nouns, as well as logogram languages such as Chinese it is unlikely to
provide a definitive solution.

An advantage of this approach is that given a pretrained Seq2Seq+Att model,
our proposed mechanism only modifies the decoding procedure, requiring
therefore no need for retraining.
The separation of the neural lexicon and a symbolic statistical lexicon may
also help in cases where certain words may have insufficient presence in
the training data, such that its occurrence at test time may drastically
increase perplexity and lead to discriminating against the right
choice of word in favour of the incorrect but more familiar choices (e.g.,
repetition, overwhelming decrease in BLEU score when presented with more than 2
\texttt{UNK} tokens on the source side).
The symbolic dictionary can be easily manipulated by humanly adjusting its
content, making explicit discouragement of such discrimination possible.

\subsection{LexPG: Dictionary Fusion with Pointer Generator}
\label{ssec:Lex:LexPG}
LexPN treats all \texttt{UNK} tokens in the neural model as placeholders for
the proposed mechanism to replace.
Such a design requires placing these placeholders at appropriate positions
during training.
This leads to a fixed notion of which OOVs exist rather than dealing with OOVs
during the NMT model training process.

We propose to use a mechanism similar to pointer generator \citep{CL2016032,
CL2017118}.
At each step, a separate neural component ($p_{gen}$, the pointer generator)
takes the current hidden states as input, and decide whether to use the neural
decoder's output or to rely on the dictionary.
Different from \citet{CL2016032}'s and \citet{CL2017118}'s approach, our proposed
models are optimised for dictionary entries and incorporate dictionary
features.
This way, the model can choose to use the dictionary even when the neural
decoder decides to output something other than \texttt{UNK}, and when using the
dictionary it may consult the neural decoder for suggestions.
We argue that this added flexibility may help handling trickier situations.
One may even hypothesise that the pointer generator can learn to trigger
depending on whether the conditional language model is confident enough with
its choice.

\textbf{LexPG}
The proposed LexPG model uses an activation unit $p_{gen}$ just like in
Equation~\ref{equ:NMT:copyPG} to decide whether the current output should be
determined by the neural decoder or the bilingual lexicon.
The detailed definitions are given in Equation~\ref{equ:Lex:LexPG}.

\begin{equation}
    \begin{aligned}
    \label{equ:Lex:LexPG}
    p_{gen} &= \sigma (w_{gen}^Tc_t + b_{gen}) \\
    \pr_{\mathrm{LexPG}}(y_t = w &| y_{<t}, X) = \\
        p_{gen} &\times \pr_{\dec}(y_t = w | y_{<t}, X) + \\
        (1 - p_{gen}) &\times \sum_{w \in \lex(x_i)} \alpha_i  \\
    \end{aligned}
\end{equation}

During inference time, the bilingual lexicon is used when $p_{gen}$ is below a
certain threshold.
One could view $p_{gen}$ as a switch between the symbolic bilingual lexicon and
the neural model.
At each step, the model takes into account all the current hidden states
as input and chooses which component output to use.

Since a source word may have multiple translations in the target language,
decisions will need to be made on which target word to use given current
context $(X, y_{<t})$.
We propose to use the conditional language model
(Equation~\ref{equ:NMT:base_dec}) as a scoring function for such purpose, as
$\pr_{\dec}(y_t|y_{<t}, X)$ is essentially a probability distribution on the
target neural lexicon.
Assuming $y_t$ is aligned to $i = \argmax(\alpha_t)$, each element $e$ within
translation candidate set $\mathrm{Cand}(y_t) = Lex(x_i) = \{e^{x_i}_0,...\}$
is scored by the conditional language model.
Just like for LexPN, the final output is fed back into the neural conditional
language model for generating the hidden state at time $t+1$.

The loss function during training is the addition of:
\begin{itemize}
    \item neural conditional language model loss (negative log loss of
        $\pr_{\dec}(y_t = w | y_{<t}, X)$ in Equation~\ref{equ:NMT:base_dec});
    \item LexPG augmented probability distribution:
        $-\log(\pr_{\mathrm{LexPG}}(y_t = w | y_{<t}, X))$;
    \item in the event of $w \not\in \mathrm{nLex}$ (the reference word for the
        neural decoder is \texttt{UNK}), then we will definitely want to
        minimise $p_{gen}$: $-\gamma\log(1 - p_{gen})$, where $\gamma=1$ iff
        $w \not\in \mathrm{nLex}$, otherwise $\gamma=0$;
\end{itemize}

\textbf{LexPG+S: Separate Pointer Scoring Layer}
More recent work points out that attention score does not equal to word
alignment, it also captures most relevant information \citep{CL2017342}.
We therefore attempt to use a separate pointer scoring function which takes the
combined decoder hidden representation and context vector as input, and
produces $\beta$ pointer scores trained to represent word alignment.

\begin{equation}
    \begin{aligned}
    \label{equ:Lex:LexPG+S_beta}
    \mathrm{score}'_{ti} &=
        V' \times \tanh(W' \times h^{\dec}_t + U' \times h^{\enc}_i) \\
    \beta &= \softmax(\mathrm{score}'_t) \\
    \end{aligned}
\end{equation}

\begin{equation}
    \begin{aligned}
    \label{equ:Lex:LexPG+S}
    p_{gen} &= \sigma (w_{gen}^Tc_t + b_{gen}) \\
    \pr_{\mathrm{LexPG+S}}(y_t = w &| y_{<t}, X) = \\
        p_{gen} &\times \pr_{\dec}(y_t = w | y_{<t}, X) + \\
        (1 - p_{gen}) &\times \sum_{w \in \lex(x_i)} \beta_i  \\
    \end{aligned}
\end{equation}

\textbf{LexPG+F: Incorporating Dictionary Entry Features}
Although the dictionary in LexPG(+S) is in theory extensible, as in it is
intuitive to add/modify entries in the dictionary post-training, the fact that
the neural activation unit $p_{gen}$ is completely agnostic to what is in the
dictionary (and what is not) may hinder its decision making capability.
We therefore add some simple dictionary entry features so that the model may
more easily leverage post-training modifications to the dictionary.
The feature generator takes a source word as input and outputs a binary vector
$f^{feat}$ containing:

\begin{itemize}
    \item whether the word exists in the dictionary;
    \item whether the word exists has a unique translation;
    \item whether the word also exists in the target side neural lexicon.
\end{itemize}

The binary vector is then mapped to dense representation using a feature
embedding matrix.
Since each source word will have different dictionary features, the score
for looking up the dictionary for each word is calculated separately using
Equation~\ref{equ:Lex:LexPG+F_PC}.

\begin{equation}
    \begin{aligned}
    \label{equ:Lex:LexPG+F_PC}
    \mathrm{PC}_i' = V^T_{PC} \tanh(&W_{PC} \times \hat{h}^{\dec} \\
                                  + &U_{PC} \times h^{\enc}_i \\
                                  + &O_{PC} \times f^{feat}_i) \\
    \end{aligned}
\end{equation}

Then, the final score for LexPG+F is computed through a softmax layer:
\begin{equation}
    \begin{aligned}
    \label{equ:Lex:LexPG+F_Final_Score}
    p_{gen}' &= V^T\tanh (W_{gen}^T\hat{h}^{\dec} + b_{gen}) \\
    p_{gen}, \mathrm{PC} &= \mathrm{softmax}([p_{gen}', \mathrm{PC}']) \\
    \pr_{\mathrm{LexPG+F}}(y_t &= w | y_{<t}, X) =\\
        p_{gen} &\times \pr_{\dec}(y_t = w | y_{<t}, X) + \\
        \sum_{\mathrm{Lex}({x_i}) \ni w} \mathrm{PC}_i &\times \alpha_i
    \end{aligned}
\end{equation}

Note since $p_{gen}$ and $\mathrm{PC}$ are the result of a single softmax
layer, we have $p_{gen} + \sum_i\mathrm{PC}_i = 1$.
Therefore the final probability calculation of
$\pr_{\mathrm{LexPG+F}}(y_t = w | y_{<t}, X)$ does not have term $(1-p_{gen})$
as in Equation~\ref{equ:Lex:LexPG} and Equation~\ref{equ:Lex:LexPG+S}.

For LexPG+S+F, we design the final score to be the combination of LexPG+S and
LexPG+F.
we discovered through experiments that a simple replacement of
$\alpha$ with $\beta $in Equation~\ref{equ:Lex:LexPG+F_Final_Score} leads to
frequent vanishing gradient.
We hypothesise that this could be caused by the lack of sufficient supervision
on $p_{gen}$.
Since all $\mathrm{PC}$ and $p_{gen}$ are passed through the same softmax
layer, and that we do not expect the proposed lexicon fusion mechanism to
trigger much more frequently than the neural generator, we take the
average of $(1-p_{gen})$ and $\mathrm{PC}$ for LexPG+S+F:

\begin{equation}
    \begin{aligned}
    \label{equ:Lex:LexPG+S+F_Final_Score}
    \pr_{\mathrm{LexPG+S+F}}(y_t &= w | y_{<t}, X) =\\
        p_{gen} &\times \pr_{\dec}(y_t = w | y_{<t}, X) + \\
        \sum_{\mathrm{Lex}({x_i}) \ni w}
            \beta_i &\times \frac{(1-p_{gen}) + \mathrm{PC}_i}{2}
    \end{aligned}
\end{equation}

\section{Experimental Results}
\label{sec:expt}
\subsection{Modelling Detail and Datasets}
We implement all models using DyNet library in Python \citep{ML2017009}.
For embedding and hidden state we use 256 dimensions, and all LSTM
\citep{ML1997001} units are stacked 2 layers.
For training, we use mini-batches of size 64.
All models are trained with Adam optimiser \citep{ML2015007} with early
stopping.

We compare the proposed models' performance on multiple language pairs.
We use German-English from IWSLT2017 and test2015.
For low-resource language pairs, we use Czech-English and Russian-English from
News Commentary v8.

\begin{table}[t!]
    \begin{center}
        \begin{tabular}{|l|l|l|l|}
            \hline
            Language pair & DE-EN     & CS-EN     & RU-EN     \\ \hline
            Train pairs   & 226,572   & 134,453   & 131,492   \\ \hline
            Test  pairs   & 1080      & 2,999     & 2,998     \\ \hline
            Uniq. src lex & 130,288   & 153,173   & 159,074   \\ \hline
            Uniq. tgt lex & 52,096    & 59,909    & 64,220    \\ \hline
            Avg. src len  & 25        & 22        & 25        \\ \hline
            Avg. tgt len  & 25        & 25        & 26        \\ \hline\hline
            Dict. size    & 171,137   & 198,627   & 207,242   \\ \hline
        \end{tabular}
    \end{center}
    \caption{\label{table:expt:datainfo} Dataset statistics.
        For dictionary generation, we use the intersection method to ensure
        high precision\citep{CL2006001}.}
\end{table}

\subsection{Main Results}
Experiments in this subsection are conducted with lexicon filtered with minimum
occurrence frequency bigger or equal to 2.
We find this setting to be fairly common in a number of literature and should
give us a fair idea on how competitive the proposed models behave without
extensive hyperparameter tuning.
Table~\ref{table:expt:main} shows the BLEU score of all compared models.

For baseline models, we compare with a standard LSTM based sequence-to-sequence
with attention (Baseline), $\texttt{UNK}$ replacement or CopyNet mechanism
(PN Copy), and pointer Generator copy mechanism (PG Copy).
The details of all these models can be found in \S\ref{sec:NMT}

\begin{table}[t!]
    \begin{center}
        \begin{tabular}{|l|l|l|l|}
            \hline
            Language pair & DE-EN & CS-EN & RU-EN \\ \hline
            Baseline      & 22.37 & 9.35  & 9.26  \\ \hline
            PN Copy       & 23.01 & 10.41 & 9.72  \\ \hline
            PG Copy       & 23.07 & 10.85 & 10.47 \\ \hline \hline
            LexPN         & 23.08 & 10.51 & 9.54  \\ \hline
            LexPG         & 23.02 & 12.25 & 10.67 \\ \hline
            LexPG+S     & \textbf{24.72} & 11.71 & 11.56 \\ \hline
            LexPG+F     & 24.38 & 12.05 & 11.24 \\ \hline
            LexPG+S+F & 24.57 & \textbf{12.49} & \textbf{12.18} \\ \hline
        \end{tabular}
    \end{center}
    \caption{
        \label{table:expt:main}
        Experimental results in BLEU.
        PN Copy stands for Pointer Network Copy mechanism (or \texttt{UNK}
        replacement as seen in some literature such as by \citet{CL2015009}).
        PG Copy stands for Pointer Generator Copy mechanism, the details of
        which can be found in \S~\ref{sec:NMT}.
    }
\end{table}

An interesting observation is that despite without much of a common lexicon,
PG Copy outperforms both PN Copy and LexPN significantly for
Russian-English and Czech-English.

Another interesting line of work is the utilisation of sub-word units such as
BPE \citep{CL2016179}.
We perform BPE on the input data and measure the BLEU score after merging
the sub-word units on the target side.
The results are shown in Table~\ref{table:expt:bpe}.

\begin{table}[t!]
    \begin{center}
        \begin{tabular}{|l|l|l|}
            \hline
            Language pair & DE-EN & DE-EN BPE \\ \hline
            Baseline      & 22.37 & 23.20     \\ \hline
            PN Copy       & 23.01 & 23.20     \\ \hline
            PG Copy       & 23.07 & 22.86     \\ \hline \hline
            LexPN         & 23.08 & 23.20     \\ \hline
            LexPG         & 23.02 & 23.57     \\ \hline
            LexPG+S     & \textbf{24.72} & 23.68     \\ \hline
            LexPG+F     & 24.38 & 24.16     \\ \hline
            LexPG+S+F & 24.57 & \textbf{24.67}     \\ \hline
        \end{tabular}
    \end{center}
    \caption{
        \label{table:expt:bpe}
        BLEU scores on the IWSLT2017 German-English dataset.
    }
\end{table}

BPE on German-English improves the baseline and PN Copy but failed for PG Copy.
Extensive experiments are conducted on different neural lexicon sizes, but
the proposed model also did not receive any significant boost, but rather seem
to underperform.
After examination of the output, we hypothesise that it might be caused by
BPE's syntactically and semantically awkward splittings such as the splittings
of proper nouns.
Although some research \citep{CL2017317} suggest that syntactically motivated
segmentation such as morphological segmentation does not perform as good as
syntax-agnostic BPE, we argue that our proposed dictionary fusion technique
may be a more meaningful and controllable alternative in handling the limits in
NMT neural lexicon.

\subsection{Coverage Study: the impact of neural lexicon size}
\label{sec:lexsize}
The proposed models contain two separate lexicons (a neural lexicon for the
neural conditional language model, and a symbolic dictionary).
While in theory, the symbolic dictionary can accommodate growing lexicon, the
neural lexicon is much less extensible and is usually size-capped.

In this section, we investigate the influence of such limit on the performance
of the proposed models.
For the sake of simplicity, we refer to the minimum number of occurrence
required to be included in the neural lexicon as $\mathrm{LexBar}$.
Statistics regarding the German-English dataset are presented in
Table~\ref{table:expt:LexBar}

\begin{table*}[t!]
    \begin{center}
        \begin{tabular}{|l|l|l|l|l|l|l|l|l|l|l|}
            \hline
            $\mathrm{LexBar} = $ & 1       & 2      & 3      & 4      & 6       & 8       & 12      & 16      & 24      & 32      \\ \hline\hline
            src. neu. lex.       & 130k    & 58k    & 41k    & 32k    & 23k     & 19k     & 14k     & 12k     & 8k      & 6.5k    \\ \hline
            src. neu. percentage & 100\%   & 45\%   & 31\%	 & 25\%   & 18\%    & 14\%    & 11\%    & 10\%    & 6\%     & 5\%     \\ \hline
            src. unk.            & 0       & 72k    & 90k    & 98k    & 107k    & 112k    & 116k    & 118k    & 122k    & 124k    \\ \hline
            src. unk. in dict.   & -       & 68\%   & 72\%   & 73\%   & 75\%    & 76\%    & 77\%    & 78\%    & 78\%    & 78\%    \\ \hline\hline
            tgt. neu. lex.       & 52k     & 34k    & 27k    & 23k    & 18k     & 15k     & 12k     & 11k     & 7.6k    & 6.3k    \\ \hline
            tgt. neu. percentage & 100\%   & 65\%   & 51\%   & 43\%   & 34\%    & 29\%    & 23\%    & 21\%    & 15\%    & 12\%    \\ \hline
            tgt. unk.            & 0       & 18k    & 25k    & 30k    & 34k     & 37k     & 40k     & 41k     & 44k     & 46k     \\ \hline
            tgt. unk. in dict.   & -       & 70\%   & 73\%   & 75\%   & 77\%    & 78\%    & 79\%    & 82\%    & 81\%    & 82\%    \\ \hline\hline
            src.unk == tgt.unk   & -       & 5k     & 7k     & 8.6k   & 10k     & 11k     & 12k     & 12k     & 13k     & 14k     \\ \hline
        \end{tabular}
    \end{center}
    \caption{
        \label{table:expt:LexBar}
        LexBar statistics of IWSLT2017 German-English dataset.
    }
\end{table*}

\begin{figure}[t]
    \centering
    \includegraphics[width=0.45\textwidth]{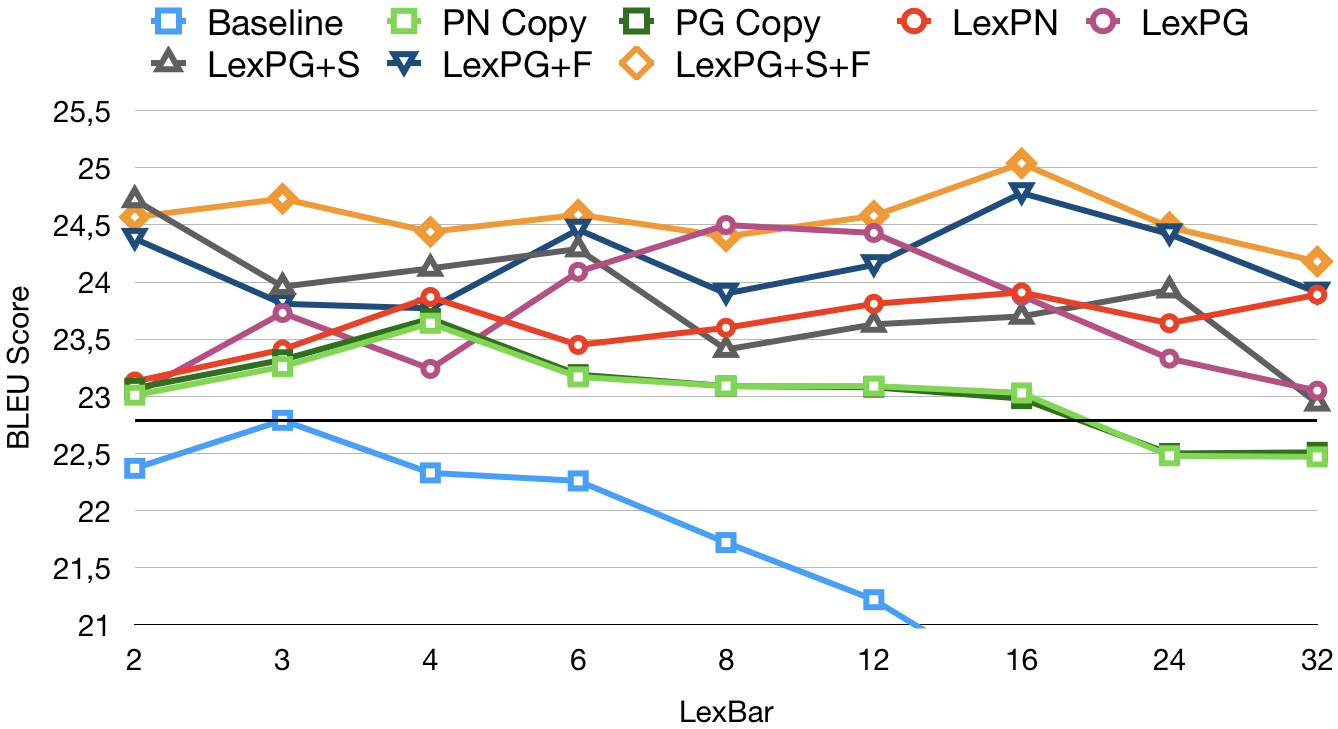}
    \caption{
        Coverage study: reduce source and target lexicon size.
    }
    \label{fig:expt:coverage1}
\end{figure}

We start with $\mathrm{LexBar} = 2$ and gradually increase it to $32$.
In this case, the Seq2Seq baseline reaches its best performance at
$\mathrm{LexBar} = 3$ then decreases more rapidly as the neural lexicon
does (Figure~\ref{fig:expt:coverage1}).

Among the proposed models, we see that LexPG's performance is significantly
better than the baseline models, but its curve is somewhat unstable.
Similar instability was also observed for Russian-English and Czech-English,
for which LexPG achieves their best BLEU scores under completely different
$\mathrm{LexBar}$ settings.
The problem with this is that given any dataset, one might need to perform
a lot of parameter search to find an optimal $\mathrm{LexBar}$ for LexPG.

Another interesting observation is that our further analysis shows that the
performance of the neural conditional language model component of LexPG
(evaluated by removing the dictionary fusion at test time) was able to benefit
from LexPG, receiving a slight boost when LexPG reaches its peak at
$\mathrm{LexBar} = 8$.
We suspect that the direct usage of attention $\alpha$ in LexPG's loss
function might have led to this outcome.

LexPG+S has a curve that decreases as $\mathrm{LexBar}$ increases, but at a
pace much more subtle than all baselines.
Comparing to LexPG, LexPG+S's $\beta$ layer was able to benefit from the rich
neural lexical coverage, although this dependency eventually led to worsened
performance as neural lexicon shrinks.

We ran additional tests and found that both LexPG and LexPG+S are much more
sensitive to source neural lexicon size than target.
This could be caused by the fact that the dictionary fusion for LexPG and
LexPG+S is not exposed to whether the source element is covered by the
dictionary or not, leaving the pointer generator to learn such coverage on its
own.
When the neural lexicon is small, LexPG(+S) will not be able to distinguish
between different unknown source words, since all \texttt{UNK} have the same
word embedding for the neural decoder.

Unlike LexPG(+S), LexPG+F is exposed to dictionary features and is so far the
most stable one: having been able to maintain overall better performance even
when the neural lexicon is reduced to around 10\%.
Where LexPG(+S)'s neural decoder may not have enough information to determine
the correct responses to source \texttt{UNK}s, LexPG+F's knowledge of the
dictionary is shown to help make better decisions.
The conflict between attention $\alpha$ and word alignments are also resolved
by the inclusion of a separate alignment $\beta$ score in LexPG+S+F, making it
our most stable and well performing model.

Since the performance of our proposed model does not drop and in some cases
improves the BLEU scores even with a smaller neural network, we think it may
also benefit production scenarios where the number of parameters are tightly
constrained (e.g., deployment on mobile devices).
In our most extreme case, LexPG's saved model (neural parameters and
pickled encoder-decoder class object with full dictionary) is almost 4 times
smaller than a baseline model with 100K (50K for source and 50K for target
side) neural dictionary (221M vs 871M).
Less neural parameters also leads to faster learning and inference.
In our experiments the training of LexPG+S+F model despite its complexity is
only 10\% slower than a baseline with the same neural lexicon.
If we compare against increased LexBar without sacrificing BLEU score,
LexPG+F's training with LexBar=16 is about 3-4 times faster than baseline
(LexBar=2).

\subsection{Expansion Study: expanding the dictionary post-training}
The purpose of this study is to evaluate the dictionary component's
extensibility.
We first train the models using the same training data above, then add more
entries into the dictionary without changing any of the neural parameters.
Theoretically it is possible to use a real bilingual dictionary, but for the
sake of simplicity, these additional entries are obtained using word alignment
of the test set.

\begin{table*}[t!]
    \begin{center}
        \begin{tabular}{|l||l|l||l|l||l|l|}
            \hline
            Language pair & DEEN  & DEEN+ & CSEN  & CSEN+ & RUEN  & RUEN+ \\ \hline
            Baseline      & 22.37 & -     & 9.35  & -     & 9.26  & -     \\ \hline
            PN Copy       & 23.01 & -     & 10.41 & -     & 9.72  & -     \\ \hline
            PG Copy       & 23.07 & -     & 10.85 & -     & 10.47 & -     \\ \hline
            \hline
            LexPN         & 23.08 & 23.24 & 10.51 & 10.99 & 9.54  & 10.08 \\ \hline
            LexPG         & 23.02 & 23.70 & 12.25 & 13.66 & 10.67 & 12.63 \\ \hline
            LexPG+S       & \textbf{24.72} & 25.30 & 11.71 & 12.79 & 11.56 & 13.28 \\ \hline
            LexPG+F       & 24.38 & 25.01 & 12.05 & 13.45 & 11.24 & 13.02 \\ \hline
            LexPG+S+F     & 24.57 & \textbf{25.50} & \textbf{12.49} & \textbf{13.89} & \textbf{12.18} & \textbf{13.58} \\ \hline
        \end{tabular}
    \end{center}
    \caption{
        \label{table:expt:expansion}
        Expansion study: adding more entries into the dictionary post-training.
        The proposed models are first trained with existing automatically
        acquired entries, then prior to testing new entries are added to
        further expand the dictionary.
    }
\end{table*}

One might think that such exposure to the test set would give our models unfair
advantage, but we argue that this is actually a noisier substitution to a
real bilingual dictionary.
Our intent here is merely to simplify dictionary collection/generation.
Ideally, a bilingual dictionary should contain the aligned components and with
better quality as well.
What we are doing is essentially adding a flawed fraction of a real bilingual
dictionary, to show how well the model can leverage such additional knowledge.
During these experiments, the neural conditional language model, its word
embedding matrices and other neural components are completely untouched.
And as the results show, our proposed models are indeed able to put these
information into good use and boost its performance.

To accommodate changes in a pretrained neural translation model, extensive data
collection and retraining is usually required for active learning (or even
life-long learning).
Our use of a bilingual lexicon provides an alternative.
The bilingual lexicon can be updated dynamically while keeping the neural model
fixed.

\subsection{Improving The Bilingual Lexicon}
At the beginning of \S\ref{sec:Lex} we claim that the dictionary entries we
used in our experiments are merely ``dictionary-like''.
Indeed, word alignment can only produce word-to-word alignments and cannot
handle phrases.
In some cases, words that should have been aligned to multi-word phrases are
only aligned to one of the words in the phrase (e.g. ``Bundeskanzler'' aligned
to ``Federal'' instead of ``Federal Chancellor'').
More over, the aligner itself is not exactly error-proof.
It is therefore reasonable to assume that there is a significant amount of
error which could be humanly corrected in the ``dictionary'' we used.

Aside from relying on alignment, there are many other potential ways to improve
the dictionary, such as through crowd-sourcing and extracting entries from a
real bilingual dictionaries.
The dictionary itself can also contain word-to-phrase translations, which can
be hugely beneficial for languages with a lot of compound words and idioms.
Through preprocessing to identify known named entities (e.g. ``Monty Python's
Life of Brian''), we may even enforce more constraints on how specific terms
should be translated.


\section{Conclusions and Future Work}
\label{sec:zukft}
We present several simple and highly practical approaches to incorporating
structured symbolic knowledge -- a bilingual dictionary -- into a standard
neural machine translation model.
Our experimental results show that these methods not only produce much better
results, but are faster to train, smaller in parameter size, and overall more
extensible to adding new knowledge.

For the future, we think it is possible to substitute the LSTM-based
neural conditional language model with a transformer \citep{CL2017244}.
Experimenting on pretraining the neural conditional language model and only
train the proposed LexPG(+S)(+F) components may also be of interest, as
this would further demonstrate the flexibility of our proposed approach.

On a separate track, we would also like to investigate the possibility of
leveraging even more structured knowledge, such as a phrase-table.
The ability to accommodate complex mapping constraints, including Synchronous
Context-Free Grammar (SCFG) is also appealing and worthy of our endeavour.

\bibliography{bib/BU,bib/CL_pre2010,bib/CL,bib/ML}
\bibliographystyle{acl_natbib}

\appendix

\end{document}